%% file: ex_article.tex
\documentclass[onefignum,onetabnum]{siamonline220329}

\input{ex_shared}

\ifpdf
\hypersetup{
  pdftitle={Learning Optimal Filters Using Variational Inference},
  pdfauthor={E. Bach, R. Baptista, E. Luk, and A. Stuart}
}
\fi

\DeclareMathOperator*{\argmin}{arg\,min}

\usepackage{color,soul}
\usepackage{mathtools}
\usepackage{todonotes}

\newcommand{\llb}{[[}
\newcommand{\rrb}{]]}
\newcommand{\lr}{\llb 1, J \rrb}

\newtheorem{dataassumption}[theorem]{Data Assumption}

\begin{document}

\maketitle

\begin{abstract}
Filtering---the task of estimating the conditional distribution for states of a dynamical system given partial and noisy observations---is important in many areas of science and engineering, including weather and climate prediction. However, the filtering distribution is generally intractable to obtain for high-dimensional, nonlinear systems. Filters used in practice, such as the ensemble Kalman filter (EnKF), provide biased probabilistic estimates for nonlinear systems and have numerous tuning parameters. Here, we present a framework for learning a parameterized analysis map---the transformation that takes samples from a forecast distribution, and combines with an observation, to update the approximate filtering distribution---using variational inference. In principle this can lead to a better approximation of the filtering distribution, and hence smaller bias. We show that this methodology can be used to learn the gain matrix, in an affine analysis map, for filtering linear and nonlinear dynamical systems; we also study the learning of inflation and localization parameters for an EnKF. The framework developed here can also be used to learn new filtering algorithms with more general forms for the analysis map.
\end{abstract}



\section{Introduction}
\label{submission}

\subsection{Overview}
\label{ssec:O}

Data assimilation (DA), the problem of estimating the state of a dynamical system given partial and noisy observations, is ubiquitous in many fields. The genesis of many DA methods was in the geosciences, and DA is now an essential component of numerical weather prediction~\cite{kalnay_atmospheric_2002} and is being used in climate prediction~\cite{carrassi_data_2018}, for example. The core task of DA, from a probabilistic perspective, is to obtain the
distribution of the time-dependent state of a dynamical system, given observations. When
this is performed sequentially, it leads to the filtering distribution.

Obtaining the filtering distribution is typically intractable for high-dimensional and nonlinear systems. In the case of linear Gaussian systems, the filtering distribution is itself Gaussian and given in closed form by the Kalman filter \cite{kalman_new_1960}; for high-dimensional dynamical systems propagating the covariance matrices can be prohibitively expensive. In the nonlinear case, the filtering distribution can be approximated by the particle filter \cite{doucet_sequential_2001}; this methodology is asymptotically consistent with the true filter, but often performs poorly in high dimensions because of weight collapse \cite{snyder_obstacles_2008}. The extended Kalman filter (ExKF)~\cite{jazwinski_stochastic_1970} 
works well in close-to-Gaussian settings in problems of moderate dimension, but is also impractical in high dimensions
because of the need to propagate covariance matrices. The ensemble Kalman filter (EnKF), introduced
by Evensen in 1994~\cite{evensen_sequential_1994} and which deploys low-rank empirical covariances, is thus used in practical high-dimensional settings.
Although the EnKF provides a consistent approximation
of the true filtering distribution for problems which are Gaussian \cite{le_gland_large_2011} or close to 
Gaussian~\cite{carrillo_mean-field_2024,calvello2025accuracy}, 
it is not consistent with the true filter in the nonlinear case.
In addition, the EnKF requires tuning parameters, such as inflation and localization parameters, to perform well in practice \cite{carrassi_data_2018}.

It is thus natural to consider \emph{learning} filters. In this paper we provide a new framework for doing so, based on variational inference. We assume that the forecast model is known and learn parameters in the analysis operator,
which combines the model forecast distribution with an observation to update the filtering distribution. We instantiate the analysis operator, which acts in the space of
probability measures, through an analysis map on state space. We use a probabilistic loss function to learn parameters in the analysis map. These parameters are defined by a Kalman filter with fixed gain, a generalized ExKF, and an EnKF; however, the framework introduced applies very generally.

The remainder of this section is organized as follows. In~\cref{ssec:SU} we provide the
mathematical set-up for probabilistic filtering. Then, \cref{ssec:L} contains a literature review which sets our work in context, describes our contributions and outlines the organization of the paper.

\subsection{Background on Filtering}
\label{ssec:SU}

We start by briefly reviewing the probabilistic formulation of the filtering problem; 
more details can be found in \cite{jazwinski_stochastic_1970,law_data_2015,majda_filtering_2012,reich_probabilistic_2015,sanz-alonso_inverse_2023}. We assume a dynamics model for evolution of the state $v^\dagger_j\in\mathbb{R}^d$ with form
\begin{subequations}
\label{eq:sdm}
\begin{align}
v^\dagger_{j+1} &= \Psi(v^\dagger_j) + \xi^\dagger_j, \quad j \in \mathbb{Z}^+, \\
v^\dagger_0 &\sim \mathcal{N}(m_0, C_0), \quad \xi^\dagger_j \sim \mathcal{N}(0, \Sigma) \quad \text{i.i.d.}\, ,
\end{align}
\end{subequations}
where the noise sequence $\{ \xi^\dagger_j \}_{j \in \mathbb{Z}^+}$ 
is independent of the initial condition $v^\dagger_0$.
The observations $y^\dagger_j\in\mathbb{R}^p$ are assumed to be given by
\begin{subequations}
\label{eq:dm}
\begin{align}
y^\dagger_{j+1} &= h(v^\dagger_{j+1}) + \eta^\dagger_{j+1} , \quad j \in \mathbb{Z}^+,\\
\eta^\dagger_{j+1} &\sim \mathcal{N}(0, \Gamma) \quad \text{i.i.d.}\, ,
\end{align}
\end{subequations}
where the sequence $\{\eta^\dagger_j\}_{j \in \mathbb{N}}$ is independent of $v^\dagger_0$ for all $j$, and  where the two sequences $\{ \xi^\dagger_j \}_{j \in \mathbb{Z}^+}$ and $\{\eta^\dagger_j\}_{j \in \mathbb{N}}$ are independent of one another.

Define $\llb 1, J \rrb:=\{1, \ldots, J\}.$
For a given and fixed integer $J$, we define $Y^\dagger_j = \{y^\dagger_1, \ldots, y^\dagger_j\}$ as the collection of observations up to any $j \in \llb 1, J \rrb$. The aim of the \emph{filtering} problem is to compute the filtering distribution $\Pi_j(v_j) = \mathbb{P}(v_j^\dagger | Y^\dagger_j)$. It can be shown that the solution to the filtering problem is defined iteratively as $j \mapsto j+1$ by operating on probability measures according to the following recursion, initialized at
$\Pi_0 = \mathcal{N}(m_0, C_0)$:
\begin{align}\label{eq:pna}
\begin{split}
&{ \text{ \bf  Prediction step:}} ~~~~\;  \widehat{\Pi}_{j+1} = \mathsf{P} \Pi_j,  \index{prediction} \\
& {\text{  \bf Analysis step:}} ~~~~~\,\,\,\,\,  \Pi_{j+1} = \mathsf{A}(\widehat{\Pi}_{j+1};y^\dagger_{j+1}). \index{analysis} 
\end{split}
\end{align}
We refer to $\mathsf{P}$ as the \emph{prediction operator} and $\mathsf{A}$ as the \emph{analysis operator}. The map $\mathsf{P}$, viewed as taking the space of probability measures\footnote{We primarily work with measures that have Lebesgue density; in this context we view $\mathsf{P}$, and other maps on probability measures, as maps acting on probability density functions.} on $\mathbb{R}{^d}$, $\mathcal{P}(\mathbb{R}^d)$, into itself, is linear and independent of time.
In contrast, the map $\mathsf{A}$ is nonlinear and depends on time through the observation $y^\dagger_{j+1}.$ The linear prediction operator $\mathsf{P}$ acts on probability density function $\pi$ according to
\begin{equation}\label{eq:pred_operator}
    \mathsf{P} \pi(v) \propto \int_{\mathbb{R}^d} \exp \left( - \frac{1}{2} \|v - \Psi(u)\|_{\Sigma}^2 \right) \pi(u) \, du;
\end{equation}
here we use the notation $\|\cdot\|^2_C \coloneqq (\cdot)^\top C^{-1}(\cdot)$ for any strictly positive and symmetric $C$.  The nonlinear analysis operator $\mathsf{A}(\cdot; y^\dagger)$ acts on
probability density function $\pi$ according to
\begin{subequations}
\label{eq:filter_bayes_inference}
\begin{align}
  \mathsf{A}(\pi;y^\dagger)(u) & = \frac{\nu_\pi(u,y^\dagger)}{\int_{\mathbb{R}^d} \nu_\pi(u,y^\dagger) \, d u},\\
  \nu_\pi(u, y) & \propto
\exp \left( - \frac{1}{2} \| y - h(u) \|_{\Gamma}^2 \right)\pi(u).
\end{align}
\end{subequations}
The analysis step multiplies a density function $\pi$ by the likelihood  $\mathbb{P}(y^\dagger_{j+1} | v_{j+1})$ for $y^\dagger_{j+1}$ under the observation model and normalizes to make the resulting probability density function integrate to 1. Thus, the analysis step in~\cref{eq:pna} corresponds to an application of Bayes' Theorem with the prior $\widehat{\Pi}_{j+1};$ this prior is often referred to as the forecast distribution.

In this paper we study filters which use exact $\mathsf{P}$ but aim to approximate $\mathsf{A}$ from
parameterized family $\mathsf{A}_{\theta}$. We learn the optimal parameter $\theta^*$ so that the filter in~\eqref{eq:pna}, with $\mathsf{A}$ replaced with the parameterized operator $\mathsf{A}_{\theta}$, is close to the true filter in~\eqref{eq:pna} arising
from using the exact analysis operator $\mathsf{A}$. 
Such parameterized analysis steps may arise, for instance, from learning a transformation that propagates the mean and covariances from the forecast to the filtering distribution as in a Kalman filter, or they may arise from parameters of a map that transform samples from the forecast to the filtering distribution. Examples of classes of such maps include an EnKF with unknown tuning parameters, or neural networks with unknown weights. The problem of learning the parameters of a parameterized analysis operator $\mathsf{A}_\theta$ acting on distributions can be posed, equivalently, as the problem of learning the parameters of a parameterized \emph{analysis map} $T_\theta(\cdot; y)$, often referred to as a transport, acting in state space, and in particular on multiple samples in state space. We will use both viewpoints in this paper, and the sample-based picture is discussed further in \cref{ssec:samples}.

\subsection{Literature Review and Contributions}
\label{ssec:L}

Chapters 9 and 10 of \cite{bach_inverse_2024} provide an overview of multiple approaches for learning parameterized filters. The bibliography in \cite{bach_inverse_2024} contains, but is not limited to, the work cited here. The first choice to be made when learning a filter is the data set to be used for training. Throughout this paper, 
we assume that time-series of observations $\{y^\dagger_j\}_{j \in \lr}$ from \cref{eq:dm} is available for this purpose, in addition to access to $\mathsf{P}$; some work uses paired time-series $\{(v^\dagger_j, y^\dagger_j)\}_{j \in \lr}$ from \cref{eq:sdm} and \cref{eq:dm} \cite{bach_inverse_2024}, while other work uses only data from \cref{eq:dm} without requiring knowledge of the forecast nor observation model \cite{sun_fuxi_2024,vaughan_aardvark_2025}. The second choice to be made when learning a filter is the objective function, and in particular whether it measures probabilistic accuracy (matching
the filtering distribution) or deterministic accuracy (matching the true state, often using mean-square error). We exclusively study the former in this paper.

Although different in focus from our work, there has been recent activity in using machine learning to train purely data-driven forecasting systems \cite{pathak_fourcastnet_2022,lam_learning_2023,bi_accurate_2023}, 
without access to knowledge of \cref{eq:sdm}. These methods are
predominantly trained for deterministic forecasting, but recent work considers the probabilistic setting \cite{price2025probabilistic}.
It is noteworthy that the data on which these methods have been trained is, in any case, itself produced using DA.

A framework for jointly learning the forecast and analysis step with a probabilistic cost function was proposed in \cite{boudier_data_2023}; however, despite learning the forecast step, the proposed methodology also required the use of paired time-series from \cref{eq:sdm} and \cref{eq:dm}. Learning filters with scoring rules, without a forecast model, was considered in \cite{brocker_probabilistic_2009}; this method was designed with a probabilistic objective. Previous approaches for variational filtering, without a focus on learning parameterized analysis maps, have been presented in \cite{marino_general_2018,campbell_online_2021}; these methods, being based on variational Bayes, are also designed with a probabilistic objective. To the best of our knowledge, our work is the first to consider the use of variational Bayes to learn the analysis step specifically, as opposed to learning both the forecast and analysis steps. In many settings it may be viewed as unwise to learn the forecast model from data when it is already known, even if only approximately. This is because it is challenging to approximate the solution operator defined  by time-dependent nonlinear partial differential equations from data \cite{li2022learning}; if an approximate forecast model is known it may be more natural to learn corrections to it \cite{levine_framework_2022}.

In some of our numerical experiments, we focus on learning a parameterized gain matrix as part of the analysis step. Previous work on learning a gain includes \cite{hoang_simple_1994,hoang_adaptive_1998,mallia-parfitt_assessing_2016,levine_framework_2022}. However, this previous work involved optimizing a mean-square error that matches the filter mean to observations, thus optimizing for state estimation and not for probabilistic estimation.
We also test here the application of our method to learning inflation and localization in an EnKF. There is a large literature on adaptively choosing inflation and localization parameters in EnKFs \cite{miyoshi_gaussian_2011,vishny_high-dimensional_2024}. However, there is no work we are aware of based on picking these parameters to match the true filter: the optimization is for state estimation.

A future application of the framework presented herein is the learning of new data assimilation algorithms using general classes of analysis maps, such as neural networks. Learning an analysis map in ensemble filters parameterized by a neural network was considered by \cite{mccabe_learning_2021}, using a mean-square error loss to match the filter mean to observations;
again, optimizing for state estimation and not for probabilistic estimation. Another work on learning a neural network analysis step is \cite{bocquet_accurate_2024}, which also used a deterministic cost function. Perhaps the paper closest to ours is~\cite{spantini2022coupling}, in which a nonlinear analysis map is learned for ensemble filtering using a probabilistic objective function; however the loss function is different from the variational Bayes-inspired loss that we consider in this paper.

The rest of the paper is organized as follows. Our framework for learning filters using variational inference is presented in~\cref{sec:vi}. Numerical experiments, illustrating the learning of gains, inflation, and localization, are presented in~\cref{sec:experiments}; our experiments are
based on a linear autonomous map, the Lorenz '96 ordinary differential equations model, and the Kuramoto--Sivashinsky partial differential equations. 
Concluding remarks are made in \cref{sec:conclusions}. 
The filtering algorithms used in the paper are reviewed in \cref{sec:filters}.

\section{Variational Inference for Parameterized Filters}\label{sec:vi}

The variational formulation of Bayes' Theorem \cite{zellner_optimal_1988,knoblauch_optimization-centric_2022} states that, given a prior $\mathbb{P}(u)$ and likelihood $\mathbb{P}(y|u)$, the posterior $\mathbb{P}(u|y)$ is obtained by solving the minimization problem
\begin{align*}
\mathsf{J}(q) &= \mathsf{D}_{\text{KL}}\bigl(q\|\mathbb{P}(u)\bigr) - \mathbb{E}^{u\sim q}[\log \mathbb{P}(y | u)],\\
\mathbb{P}(u|y) &= \argmin_{q \in \mathcal{P}(\mathbb{R}^d)} \mathsf{J}(q);
\end{align*}
here $\mathcal{P}(\mathbb{R}^d)$ is the space of probability measures on $\mathbb{R}^d$, $\mathbb{E}^{u\sim q}$ denotes expectation under $u$ distributed according to $q$, and $\mathsf{D}_{\text{KL}}$ is the Kullback--Leibler (KL) divergence. This divergence is defined, for two measures with probability density functions $q$ and $\pi$, by
\begin{equation} \label{eq:KLdivergence}
    \mathsf{D}_\text{KL}(q\|\pi) = \int \log \left(\frac{q(u)}{\pi(u)}\right) q(u) \, du.
\end{equation}

We now recall that the analysis step in \cref{eq:pna} is an application of Bayes' Theorem with likelihood $\mathbb{P}(y^\dagger_{j+1} | \cdot)$ and prior $\widehat{\Pi}_{j+1}(\cdot)$. The variational formulation of Bayes' Theorem can be used to transform this Bayesian inference problem to the solution of the optimization problem 
\begin{subequations}
\label{eq:var_filtering}
\begin{align}
\mathsf{J}_{j+1}(q) &= \mathsf{D}_{\text{KL}}(q\|\widehat{\Pi}_{j+1}) - \mathbb{E}^{v\sim q}[\log \mathbb{P}(y^\dagger_{j+1} | v)],\\
\Pi_{j+1} &= \argmin_{q \in \mathcal{P}(\mathbb{R}^d)} \mathsf{J}_{j+1}(q).
\end{align}
\end{subequations}
Solution of this optimization problem defines a map from measure $\widehat{\Pi}_{j+1}$ to
measure $\Pi_{j+1}$, parameterized by data $y^\dagger_{j+1}.$ The map corresponds to the analysis operator $\mathsf{A}$ defined in \cref{eq:pna}.

To approximate $\mathsf{A}$ we use the variational formulation \cref{eq:var_filtering} as a building block to define our objective function, but we optimize over parameterized subsets of the space of all probability measures. In \cref{ssec:offline} we introduce offline formulations of this problem, and in \cref{ssec:online} we modify these ideas to accommodate the online setting where the a map is learned at each new analysis step in time. \Cref{ssec:samples} is devoted to a sample-based reformulation of the learning objective, which is better suited for learning ensemble-based filtering algorithms.

Throughout this work we make the following assumption on the data available for learning.
\begin{dataassumption}
    \label{da:da} For some fixed integer $J$, the data $Y_J^\dagger$ is given. Furthermore,
    we are able to sample from the Markov transition kernel $\mathsf{P}$: given any $u \in \mathbb{R}^d$ we can draw the random variable $v$ defined by
    \begin{equation}\label{eq:sample_state}
        v = \Psi(u) + \xi, \quad  \xi \sim \mathcal{N}(0, \Sigma);
    \end{equation}
    and, given any $v \in \mathbb{R}^d$, we are able to evaluate the likelihood of $y|v$ defined by
    $$y = h(v) + \eta, \quad  \eta \sim \mathcal{N}(0, \Gamma).$$
\end{dataassumption}

\subsection{Offline Formulations}\label{ssec:offline}

Recall that the prediction $\mathsf{P}$ and analysis $\mathsf{A}(\cdot;y^\dagger_{j+1})$ operators
define the filtering update \cref{eq:pna}. Now assume that we do not have access to the true analysis operator $\mathsf{A}$, rather a $\theta$-parameterized family of analysis operators 
$\mathsf{A}_\theta(\cdot; y)$, defined for any $y \in \mathbb{R}^p$.
We may then define an optimization problem over $\theta$ rather than an optimization problem over densities. Mirroring the prediction--analysis cycle in~\cref{eq:pna}, we first define a recursion for densities $q_j$ that approximate the filtering distribution:
\begin{align}\label{eq:recursion}
    q_{j+1}(\theta) &= \mathsf{A}_\theta(\mathsf{P}q_j(\theta); y^\dagger_{j+1}),\quad q_0 = \Pi_0.
\end{align}
We then consider the optimization problem\footnote{Abusing notation by using the same symbol for the
objective function over parameters as used over probability measures previously.}
\begin{subequations}\label{eq:offline}
    \begin{align}
    \mathsf{J}_{j+1}(\theta) &= \mathsf{D}_{\text{KL}}\bigl(q_{j+1}(\theta)\|\mathsf{P}q_j(\theta)\bigr) - \mathbb{E}^{v_{j+1}\sim q_{j+1}(\theta)}[\log \mathbb{P}(y^\dagger_{j+1} | v_{j+1})],\label{eq:obj_offline}\\
    \mathsf{J}(\theta) &= \sum_{j=1}^{J} \mathsf{J}_{j}(\theta), \label{eq:sum_objs}\\ 
    \theta^* &= \argmin_{\theta} \mathsf{J}(\theta).
\end{align}
\end{subequations}
Assuming that there is some $\theta^*$ such that $\mathsf{A}_{\theta^*}(\cdot;y) = \mathsf{A}(\cdot;y)$ for all data $y$, then the analysis operator will define the true filtering distribution, i.e.,  
$\Pi_{j+1} = \mathbb{P}(v^\dagger_{j+1}|Y^\dagger_{j+1}) 
= \mathsf{A}_{\theta^*}(\widehat{\Pi}_{j+1};y_{j+1}^\dagger)$ for all $1 \leq j \leq  J$.
Given that $\mathsf{J}_{j+1}(\cdot)$ is minimized at the true filtering distribution at every $j$, $\mathsf{J}(\cdot)$ is also minimized by parameter $\theta^*$.  

In practice, however, we expect only to obtain an approximation of the true filter, primarily because the $\theta$ parameterization of the analysis operator will typically not be sufficiently rich; furthermore the global minimum $\theta^*$ of the objective function may not be found. To implement the optimization over $\theta$, under~\Cref{da:da},
we will use Monte Carlo approximations in the evaluation of the objective function. The parameterized analysis operators that we consider arise from two classes of filtering algorithms: 1) Gaussian filters; and 2) ensemble Kalman filters. In the first case, the approximate distribution $q_{j}$ will be Gaussian at each step, as will $\mathsf{P} q_j$, so the KL divergence term in the loss function $\mathsf{J}_{j+1}$ can be evaluated analytically. Moreover, the expectation of the log-likelihood can be evaluated analytically for a linear observation model $h$ and using
Monte Carlo in general; we do the latter here. In the second case, $\mathsf{A}_\theta$ is defined through a transport map, facilitating approximation of the likelihood contribution using ensembles, also a form of Monte Carlo
approximation; details are given in~\cref{ssec:samples}. To implement the variational objective in the
second case, the computation of the KL divergence will require density evaluations; 
how we do this this will also be detailed in~\cref{ssec:samples}.

\begin{remark}
    Implicit in this methodology is the idea that it will be used in a setting where $J$ is sufficiently large so that the optimal $\theta^*$ is not tuned to a specific realization
    of data $Y_J^\dagger$, but rather to the stationary statistics of the combined dynamics and observation process defined by \cref{eq:sdm}, \cref{eq:dm}. The DA method with that chosen $\theta^*$ can then be used to assimilate different data realizations derived from the same model.
    It should thus be viewed as an \emph{offline} methodology for this reason. An alternative to using large $J$ is to use multiple data realizations from \cref{eq:sdm}, \cref{eq:dm}. Our variational optimization approach is readily generalized to this setting.
\end{remark}

\begin{remark}
    The framework presented herein can be used not only for learning parameterized filters, but also for evaluating the performance of a filtering methodology, by using $\mathsf{J}$ as an evaluation metric.
\end{remark}

\subsection{Online Formulations}\label{ssec:online}

Rather than seeking a time-independent analysis map $\mathsf{A}_\theta$, we may introduce an
\emph{online} approach. We consider time-dependent parameters by introducing a new recursion for a density $q^*_j$ and learning a new parameter $\theta_j$ for every time step:
\begin{subequations}\label{eq:online}
    \begin{align}
    q^*_{j+1}(\theta) &= \mathsf{A}_{\theta}(\mathsf{P}q^*_j(\theta_j^*); y^\dagger_{j+1}),\quad q^*_0 = \Pi_0,\\
    \mathsf{J}_{j+1}(\theta) &= \mathsf{D}_{\text{KL}}\bigl(q^*_{j+1}(\theta)\|\mathsf{P}q^*_j(\theta_j^*)\bigr) - \mathbb{E}^{v_{j+1}\sim q^*_{j+1}(\theta)}[\log \mathbb{P}(y^\dagger_{j+1} | v_{j+1})],\label{eq:online_cost}\\
    \theta^*_{j+1} &= \argmin_{\theta} \mathsf{J}_{j+1}(\theta).\label{eq:theta_star2}
\end{align}
\end{subequations}
Again, if the parameterization is sufficiently rich, then the true filter can be recovered
at the optimum, but in general optimization over $\theta$ will yield only an approximation to the true filter, with accuracy limited by the expressive capability of the parameterized model.

\begin{remark}
    The online methodology may overfit to the specific data realization $Y_J^\dagger.$ However, if the resulting parameters $\theta^*_j$ are averaged over $j$ then this can be used for other data realizations, leading to an offline methodology given by 
    $$\theta^*=\frac{1}{J}\sum_{j=1}^J \theta^*_j.$$
    Note that this choice of $\theta^*$ will typically be suboptimal in comparison with the methodology from the preceding subsection. However, it will typically be cheaper to compute because the objective function only evaluates the approximation to the filtering distribution at one time.
\end{remark}

\subsection{Sample-Based Formulation}\label{ssec:samples}

In this subsection we consider an alternative viewpoint to learning parameterized filters that is natural for measures given as samples, such as those that appear in particle-based/ensemble algorithms. In this setting we learn an operator on measures by representing its action as the pushforward under a transport map $T_\theta(\cdot; y) \colon \mathbb{R}^d  \rightarrow \mathbb{R}^d$ depending on any observation $y \in \mathbb{R}^p$; that is, defining $\mathsf{A}_\theta(\pi;y)=T_\theta(\cdot; y)_\# \pi$. We then modify recursion \cref{eq:recursion} to become
\begin{align*}
    q_{j+1}(\theta) &= T_\theta(\cdot; y^\dagger_{j+1})_\#(\mathsf{P}q_j)(\theta),\quad q_0 = \Pi_0.
\end{align*}
Suppose that the $q_j$ are empirical measures consisting of $N$ samples, i.e.,
\begin{equation*}
    q_j = \frac{1}{N} \sum_{n=1}^N \delta_{v_j^{(n)}},
\end{equation*}
where $\delta_x$ is a Dirac delta function centered at $x$. Then, $q_{j+1}$ is also an empirical measure whose samples are defined by first sampling from the forecast model according to \cref{eq:sample_state}, i.e., $\widehat{v}_{j+1}^{(n)} = \Psi(v_{j}^{(n)}) + \xi_j^{(n)}$, followed by evaluating the transport map. That is,
\begin{equation*}
    q_{j+1} = \frac{1}{N} \sum_{n=1}^N \delta_{v_{j+1}^{(n)}}, \qquad v_{j+1}^{(n)} = T_\theta(\widehat{v}_{j+1}^{(n)}; y^\dagger_{j+1}).
\end{equation*}
To simplify notation we have suppressed the fact that the transport map 
$T_\theta(\cdot; y)$ will typically depend additionally on the collection
$\{\widehat{v}_{j+1}^{(n)}\}_{n=1}^N$, but in a permutation-invariant way. The reader
may consult \cref{sec:filters} for details of such dependence for the EnKF.

The log-likelihood term in \cref{eq:obj_offline} then becomes
\begin{equation*}
    -\mathbb{E}^{v_{j+1}\sim q_{j+1}(\theta)}[\log \mathbb{P}(y^\dagger_{j+1} | v_{j+1})] = -\frac{1}{N}\sum_{n=1}^N\log \mathbb{P}(y^\dagger_{j+1} | v^{(n)}_{j+1}).
\end{equation*}
As mentioned above, the filters that we optimize over are of two types: Gaussian filters and ensemble Kalman filters. In the second case the KL divergence term in~\cref{eq:obj_offline} cannot be directly computed as the divergence is not defined when the arguments are empirical measures. This requires sample-based estimators of the KL divergence such as those proposed in \cite{perez-cruz_kullback-leibler_2008,wang_divergence_2009}, or  density estimation for both arguments of the divergence in order to evaluate the integrand in~\eqref{eq:KLdivergence}. Given that sample estimators have large sample complexity in high-dimensional settings, we consider the latter approach in this work.

To explain how we use density estimators, we introduce $\mathcal{D}: \mathbb{R}^{d\times N}\to\mathcal{P}(\mathbb{R}^d)$ to be a mapping from an ensemble of size $N$ in 
$\mathbb{R}^{d}$ to a probability measure with property that 
$$\mathcal{D}\left(\frac{1}{N}\sum_{n=1}^N v^{(n)}\right) \approx q,$$ 
when $v^{(n)}\sim q$. We can then make the replacement 
$$\mathsf{D}_{\text{KL}}(q_{j+1}(\theta)\|\mathsf{P}q_j(\theta)) \mapsto
\mathsf{D}_{\text{KL}}\Bigl(\mathcal{D}(q_{j+1}(\theta))\|\mathcal{D}\bigl(\mathsf{P}q_j(\theta)\bigr)\Bigr).$$ 
In the implementations below, we take $\mathcal{D}$ to be a Gaussian fit given by computing the empirical mean and empirical covariance matrix of the ensemble; this is sometimes referred to as Gaussian projection~\cite{calvello_ensemble_2024} or as the maximum likelihood estimator of the samples over Gaussian distributions. The choice of Gaussian projections facilitates the analytical computation of the divergence. The KL divergence between two Gaussian distributions in $\mathbb{R}^d$ is given by
\begin{equation}\label{eq:kl_gaussian}
    \mathsf{D}_\text{KL}\bigl(\mathcal{N}(m_1, C_1)\|\mathcal{N}(m_2, C_2)\bigr) = \frac{1}{2}\left[\log\frac{\det(C_2)}{\det(C_1)} - d + \operatorname{tr}(C_2^{-1}C_1) + \|m_2 - m_1\|^2_{C_2}\right].
\end{equation}

\begin{remark} \label{rem:to}
Transport approaches can also be adopted in the online formulation \cref{eq:online_cost}.
There are some further simplifications of the foregoing   
that arise in this context; these simplifications
avoid use of one of the density estimations.
From the invariance of KL under invertible transformations (see, e.g., Chapter 12 of \cite{bach_inverse_2024}), and the change of variables formula for a probability density function, we have that
\begin{align*}
    \mathsf{D}_\text{KL}(q_{j+1}^*\|\mathsf{P}q_j^*) &= \mathsf{D}_\text{KL}\bigl(\mathsf{P}q_j^*\|T_\theta(\cdot; y^\dagger_{j+1})^\#\mathsf{P}q_j^*\bigr),\\
    &= \mathbb{E}^{v\sim\mathsf{P}q_j^*}[\log \mathsf{P}q_j^*] - \mathbb{E}^{v\sim\mathsf{P}q_j^*}\left[\log \mathsf{P}q_j^*\bigl(T_\theta(v; y^\dagger_{j+1})\bigr) - \log\det\left|\nabla_v T_\theta(v; y^\dagger_{j+1})\right|\right].
\end{align*}
Since, in the online algorithm, the first term does not depend on $\theta$, we can redefine
\begin{subequations}
\begin{align}
    \mathsf{J}_{j+1}(\theta) &= -\mathbb{E}^{v\sim\mathsf{P}q_j^*}\left[\log \mathsf{P}q_j^*\bigl(T_\theta(v; y^\dagger_{j+1})\bigr) - \log\det\left|\nabla_v T_\theta(v; y^\dagger_{j+1})\right|\right]\\
    &\qquad- \mathbb{E}^{v_{j+1}\sim q^*_{j+1}(\theta)}\left[\log \mathbb{P}(y^\dagger_{j+1} | v_{j+1})\right],\nonumber\\
    &\approx -\frac{1}{N}\sum_{n=1}^N\left[\log \mathcal{D}(\mathsf{P}q_j^*)\bigl(T_\theta(\widehat{v}^{(n)}_j; y^\dagger_{j+1})\bigr) - \log\det\left|\nabla_v T_\theta(v; y^\dagger_{j+1}) \right|\Bigg|_{v=\widehat{v}^{(n)}_j}\right]\\
    &\qquad- \frac{1}{N}\sum_{n=1}^N\log \mathbb{P}(y^\dagger_{j+1} | v^{(n)}_{j+1}),\nonumber
\end{align}
where $\widehat{v}_j^{(n)}\sim \mathsf{P}q^*_j$. Note that operator $\mathcal{D}$ is
applied only once, to $\mathsf{P}q_j^*$.
\end{subequations}
\end{remark}

\section{Numerical Experiments}\label{sec:experiments}

We carry out several numerical experiments for learning parameters in approximate filters: learning a gain for filtering linear and nonlinear dynamical systems, and learning inflation and localization parameters for an EnKF. Unless otherwise stated, we use the offline method in our experiments. \Cref{ssec:model} introduces the forecast and observation 
model problems that are our focus: a linear-Gaussian model, the Lorenz '96 model, and the Kuramoto--Sivashinsky equation. In \cref{ssec:gain} we learn the gain
for the linear model and in \cref{ssec:gain2} for the Lorenz '96 model. \Cref{ssec:I&L} is devoted to learning inflation and localization; both the Lorenz '96 model and the
Kuramoto--Sivashinsky equation are considered. Before proceeding to these model problems and numerical
examples, we summarize some implementation details relating to the optimization.

\subsection{Implementation details} Automatic differentiation is used to optimize both the offline and online cost functions. The expectations are approximated as empirical means using Monte Carlo samples. 
 In the offline method,  objective function~\cref{eq:sum_objs} is minimized. In so doing, the optimization uses the entire observation set $Y^\dagger_J$ to determine the filter parameters, finding the parameters that achieve the optimal filter ``on average'' over the time window. 
 
 Obtaining the gradient of the objective function over the entire trajectory---computing the sensitivity of the error in the filtering distribution at each assimilation time in the sum~\cref{eq:sum_objs}---can be expensive because of the need to differentiate through many compositions of the prediction and analysis steps; however, the computational cost is amortized in the sense that the learned parameters can be used for future filtering at no additional computational cost, which can offset this initial investment. The offline method can be considered a type of amortized variational inference \cite{marino_general_2018}. In the online method, the cost function \cref{eq:online_cost} is optimized at each time $j$, resulting in time-dependent parameters $\theta_j^*$. Thus the gradient only needs to be taken over one analysis step at a time and not the forecast step, reducing the complexity of the automatic differentiation component of the optimization methodology. In addition, one of the density estimations can be avoided in the online algorithm, as discussed in \cref{rem:to}.
    The online method can also work for a time-dependent dynamics model $(\Psi_j, \Sigma_j)$ or observation model $(h_j, \Gamma_j)$, in a straightforward way. The offline method can also be
    adapted to this setting, provided that the transport map $T_\theta$ is constructed carefully to reflect the time-dependent nature of the data acquisition process.

    We use {\tt jax} \cite{jax2018github} for automatic differentiation and gradient descent to optimize the cost functions. The open-source code for the experiments in this paper 
    is available at
\url{www.github.com/enochcluk/variational_filtering}.

\subsection{Model Problems} \label{ssec:model}

\paragraph{Linear Model} A matrix $A$ that defines a linear forecast dynamics model $\Psi(\cdot) = A\cdot$ was randomly generated as follows. First, a $40\times 40$ random matrix $W$ was created with i.i.d.\thinspace entries drawn from a standard Gaussian, i.e., $W_{ij} \sim \mathcal{N}(0,1)$ for $1 \leq i,j \leq 40$. Then, the matrix was symmetrized by taking the average with its transpose: $W'=\frac12(W + W^\top)$. Finally matrix stability was imposed: ensuring all eigenvalues are inside the unit circle.
This stability is achieved by diagonalizing $W'$, rescaling the eigenvalues  by $\lambda_\text{max} + \frac{1}{10}$, where $\lambda_\text{max}$ is the maximum  of the moduli of the eigenvalues of $W'$; and then a new matrix is reconstituted from these rescaled eigenvalues using the 
same change of basis defining the original diagonalization of $W'$.
The resulting stable matrix is $A$.

The process noise covariance matrix, $\Sigma$, is defined as $\Sigma = QQ^\top + \frac{1}{10}I$, with $Q$ being a $40\times 40$ matrix with i.i.d.\thinspace entries distributed as $\mathcal{N}(0, 0.25)$. The observation matrix and noise covariance matrix, $H$ and $\Gamma$, were set to the identity, and the initial state mean, $m_0$, to be a vector of ones. Observations were made at every time step.

\paragraph{The Lorenz '96 Atmospheric Model} The Lorenz '96 model \cite{lorenz_predictability:_1996} is a model of the midlatitude atmosphere, with chaotic response in the parameter regime we deploy, and is widely used to test data assimilation algorithms \cite{reich_probabilistic_2015}. It is given by
\begin{equation}
\frac{dx_i}{dt} = -x_{i-1}(x_{i-2} + x_{i+1}) - x_i + F,
\end{equation}
where the indices $i \in \llb 1, D\rrb $, and periodicity is used to define indices outside this set. All our experiments employ $D = 40$ and $F = 8$. The state $v$ here is $[x_1,\ldots,x_D]^\top$.

The model is integrated with fourth-order Runge--Kutta and a time step of 0.05, and observations assimilated every time step. As in the linear problem, we use $H = I$ and $\Gamma = I$. We employ the diagonal process noise covariance matrix $\Sigma = \frac{1}{10}I$.

\paragraph{The Kuramoto--Sivashinsky Model} The Kuramoto--Sivashinsky model is a one-dimensional chaotic partial differential equation, initially derived to describe flame propagation, but arising
more generally as an amplitude equation in many settings. It is given by the PDE
\begin{equation}
    u_t + u_{xxxx} + u_{xx} + u u_x = 0 , \quad (x,t) \in \mathbb{S}_L \times (0,T),
\end{equation}
where $\mathbb{S}_L$ is the circle of length $L.$
We use $L = 22$ in all reported experiments. We integrate the model using a pseudospectral method and exponential time differencing fourth-order Runge--Kutta \cite{kassam_fourth-order_2005} with $D = 256$ and a time step of $\frac14$. We use a uniform spatial grid and the state $v$ corresponds to the approximate solution at these nodes. Observations were made every 5 time steps, with $H = I$ and $\Gamma = \frac12 I$. We employ the process noise covariance matrix $\Sigma = \frac{1}{100}I$.

\subsection{Gain Learning for the Stable Linear Dynamical System}
\label{ssec:gain}

Here, we assume that $\Psi(\cdot) = A\cdot$ and $h(\cdot) = H\cdot$. We define the following recursion for the mean $m_j$ and covariance $C_j$ of a Gaussian filter which is similar to the Kalman filter, but with a frozen gain matrix $K$ that does not evolve:
    \begin{subequations}\label{eq:kalman_fixed_cov}
        \begin{align}
            \widehat{m}_{j+1} &= Am_{j},\quad \widehat{C}_{j + 1} = A C_{j} A^\top + \Sigma,\label{eq:kalman_fixed_cov_pred}\\
            m_{j+1} &= \widehat{m}_{j+1} + K(y^\dagger_{j+1} - H\widehat{m}_{j+1}),\label{eq:kalman_fixed_cov_mean_ana}\\
                        C_{j+1} &= (I - KH)\widehat{C}_{j+1}(I-KH)^\top + K\Gamma K^\top.\label{eq:kalman_fixed_cov_cov_ana}
        \end{align}
    \end{subequations}
See \cref{sec:gain_appendix} for more details on this filter. We will attempt to learn $\theta := K$. To link to our abstract formulation note that \cref{eq:kalman_fixed_cov_pred} corresponds to application of  $\mathsf{P}$ to 
$\Pi_j=\mathcal{N}(m_j,C_j)$ to find $\widehat{\Pi}_{j+1}=\mathcal{N}(\widehat{m}_{j+1},\widehat{C}_{j+1}).$ On the other hand \cref{eq:kalman_fixed_cov_mean_ana} and \cref{eq:kalman_fixed_cov_cov_ana} corresponds to application of $\mathsf{A}_\theta(\cdot;y^\dagger_{j+1})$
to map $\widehat{\Pi}_{j+1}$ into $\Pi_{j+1}=\mathcal{N}(m_j,C_j).$ We write $K_{\mathsf{opt}}$ for the learned $\theta^*.$

Experiments were conducted over $J = 1000$ steps. We used $N=10$ Monte Carlo samples for approximating the expectations of the log-likelihood. The learning rate for gradient descent was set to $\alpha = 10^{-5}$ over 100 iterations, which we found to be sufficient for the cost function to no longer significantly decrease.

We consider assimilation with full observation of the state in \cref{sssec:offline_full}. We also test partial observations in \cref{sssec:offline_partial}, wherein only every other state variable is observed. In \cref{sssec:online} we extend our
experiments to the online method.

\subsubsection{Offline Experiments: Full Observations}\label{sssec:offline_full}

 We first test learning a fixed gain $K$ using the offline method. With the exact Kalman filter solution, with varying gain $K_j$, available as a baseline, we can validate our methodology and evaluate its error. We also note that under stability conditions on the system that we satisfy, the Kalman gain $K_j$ will approach a steady-state gain $K_{\text{steady}}$ as $j\to\infty$, and the covariance matrix will also approach a steady-state $C_\text{steady}$; see \cref{sec:kf}. We compare our approximated constant gain to the steady-state gain.

\begin{figure}
    \centering
    \begin{subfigure}[t]{\linewidth}
    \centering
        \includegraphics[width=0.7\linewidth]{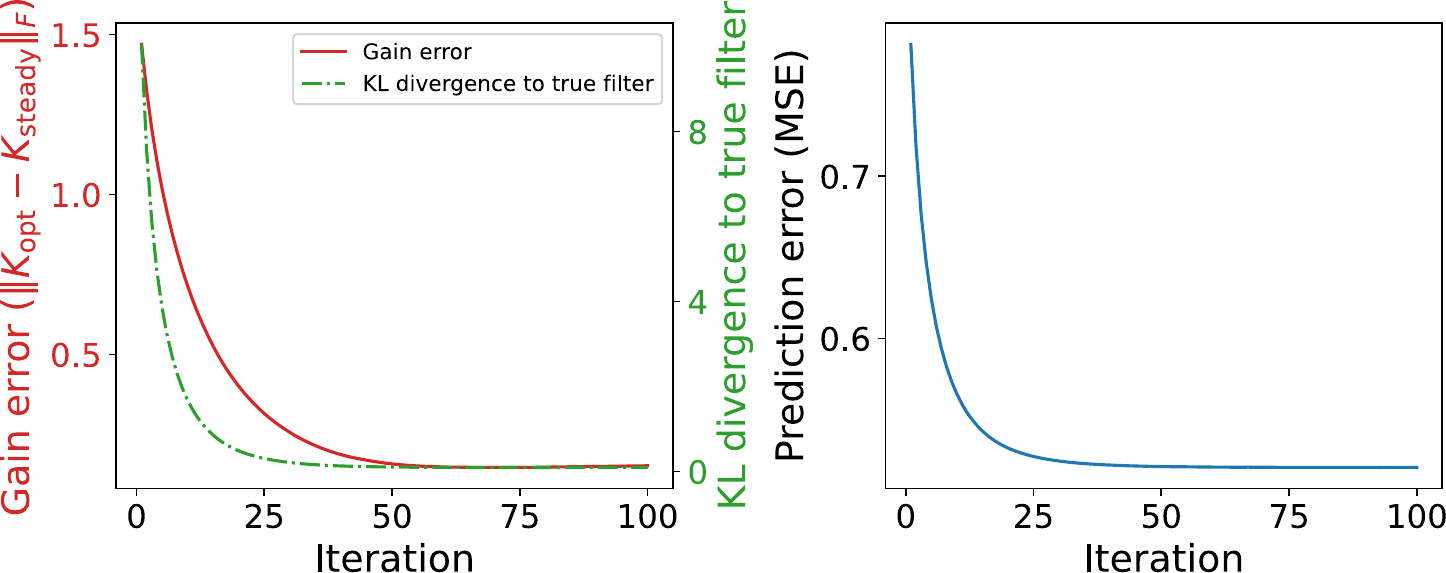}
        \caption{Errors as a function of gradient descent iterations. The Frobenius norm of the difference between the learned gain and the steady-state gain (red solid line). The KL divergence between the learned filtering distribution and the Kalman filter distribution (green dashed line). The mean-square prediction error in the filter mean compared to the true trajectory (blue solid line).}
        \label{sfig:linear_gain_errs}
    \end{subfigure}\hfill
    \begin{subfigure}[t]{\linewidth}
        \centering
        \includegraphics[width=0.7\linewidth]{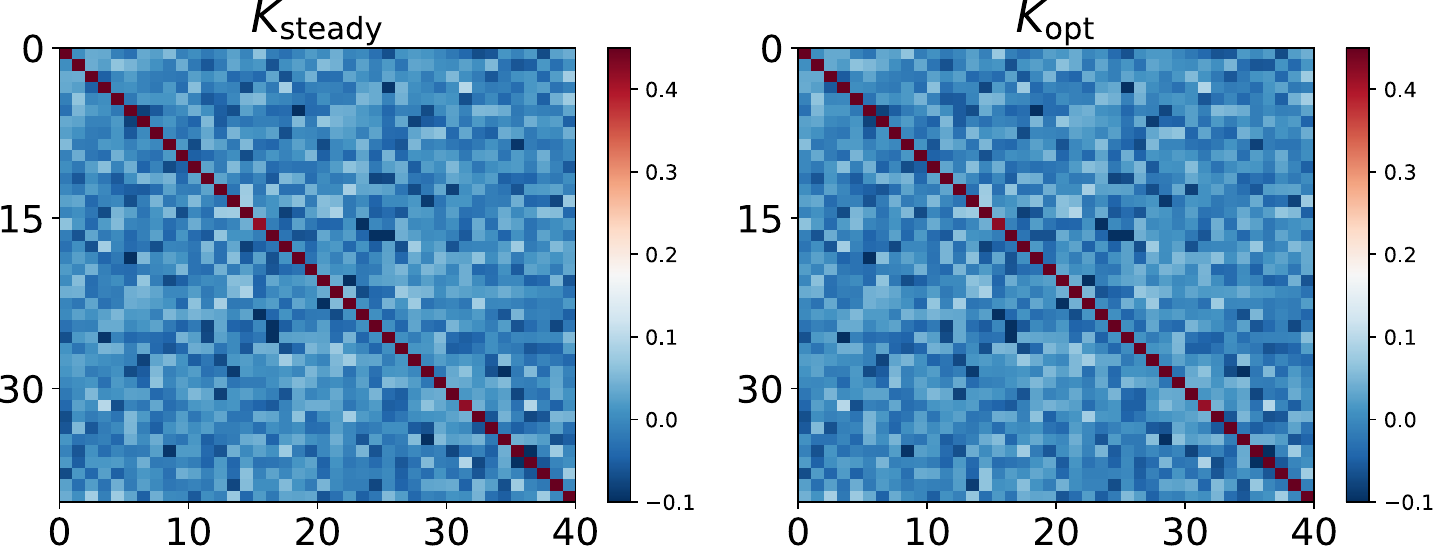}
        \caption{A comparison between the steady-state Kalman gain (left) and the learned gain (right).}
        \label{sfig:linear_gains}
    \end{subfigure}
    \caption{Learning the gain matrix for a linear dynamical system with full state observations.}
    \label{fig:linear_gain}
\end{figure}


\Cref{sfig:linear_gain_errs} shows the KL divergence between the learned filter and the true filter, the norm of the difference between the learned gain and the steady-state gain, and the mean-square prediction error of the state, as a function of the
iteration in training. All of these error measures decrease with more gradient descent steps. \Cref{sfig:linear_gains} shows a visual comparison between the steady-state Kalman gain and the learned gain.

\subsubsection{Offline Experiments: Partial Observations}\label{sssec:offline_partial}

\begin{figure}
    \centering
    \includegraphics[width=0.7\linewidth]{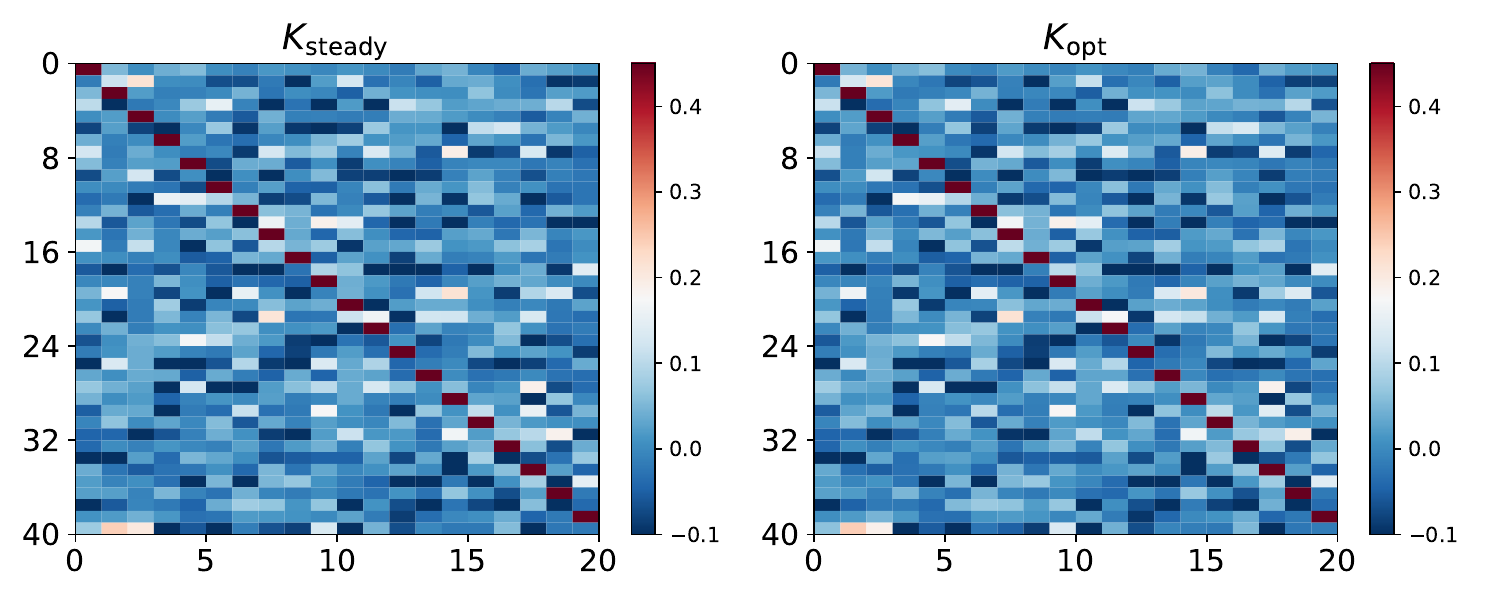}
    \caption{A comparison between the steady-state Kalman gain (left) and the learned gain (right) with partial observations.}
    \label{fig:partial_obs}
\end{figure}

In order to study the effect of partial observations, the observation matrix $H$ is modified to observe every second state variable by sub-selecting every other row of the original matrix $H$. The resulting algorithm for finding the parameters $\theta$ in converged in approximately 500 gradient descent iterations. \Cref{fig:partial_obs} shows a faithful recovery of the steady-state Kalman gain by this method.

\subsubsection{Online Experiments}\label{sssec:online}

\begin{figure}
    \centering
    \includegraphics[width=0.7\linewidth]{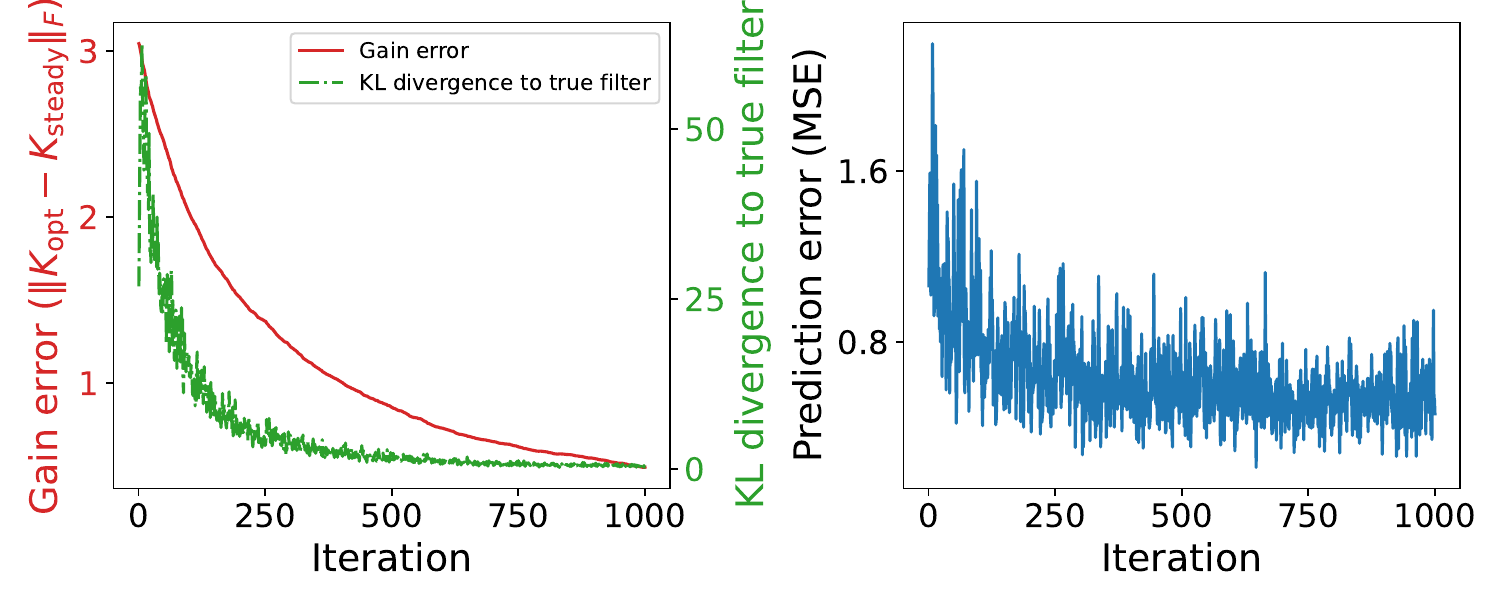}
    \caption{Errors as a function of time step for learning the gain with the online method. Within each time step, 100 gradient descent steps are performed. The Frobenius norm of the difference between the learned gain and the steady-state gain (red solid line). The KL divergence between the learned filtering distribution and the Kalman filter distribution (green dashed line). The mean-square prediction error in the filter mean compared to the true trajectory (blue solid line).}
    \label{fig:linear_gain_online}
\end{figure}
Here we employ 
100 gradient descent steps at each time step to identify the optimal parameters, for $J = 1000$ time steps. The learning rate was set to $\alpha = 10^{-5}$.
\Cref{fig:linear_gain_online} shows the results for learning the gain. Comparing to \cref{fig:linear_gain}, we can again see convergence towards the steady-state gain. However, the results are not directly comparable since the offline case shows an epoch (pass through the entire dataset) for each iteration, while the online case only involves a single epoch over 1000 time steps.

\subsection{Gain Learning for the Lorenz '96 Model}
\label{ssec:gain2}

Here we define $\Psi$ to be the solution operator for the Lorenz '96 model over 0.05
time units. To define the algorithm class within which we seek to find an optimal
filter, we let $J_j$ denote the Jacobian of $\Psi$ evaluated at $m_j$. We then
propose to learn $K$ in the following modification of algorithm \cref{eq:kalman_fixed_cov}:
\begin{subequations}\label{eq:kalman_fixed_cov2}
        \begin{align}
            \widehat{m}_{j+1} &= \Psi(m_{j}),\quad \widehat{C}_{j + 1} = J_j C_{j} J_j^\top + \Sigma,\label{eq:kalman_fixed_cov2_pred}\\
            m_{j+1} &= \widehat{m}_{j+1} + K(y^\dagger_{j+1} - H\widehat{m}_{j+1}),\label{eq:kalman_fixed_cov2_mean_ana}\\
                        C_{j+1} &= (I - KH)\widehat{C}_{j+1}(I-KH)^\top + K\Gamma K^\top.\label{eq:kalman_fixed_cov2_cov_ana}
        \end{align}
    \end{subequations}
We note that \cref{eq:kalman_fixed_cov2_pred} corresponds to an approximation
of $\mathsf{P}$ from \cref{eq:pna}. Steps \cref{eq:kalman_fixed_cov2_mean_ana} and
\cref{eq:kalman_fixed_cov2_cov_ana} correspond to an approximation
of $\mathsf{A}$ from \cref{eq:pna}. We set $\theta=K$, as in the linear case, and denote the approximation by $\mathsf{A}_\theta.$  A similar linearization to propagate covariances is also made in the extended Kalman filter \cite{jazwinski_stochastic_1970}; the method derived also has the form of cycled 3DVar \cite{lorenc_analysis_1986} with added covariance propagation 
\cite{sanz-alonso_inverse_2023}. As in the linear problem in the preceding subsection, 
we are approximating the filtering distribution at time $j$ by the Gaussian $\mathcal{N}(m_j(K), C_j(K))$. We use $J=1000$ time steps, and a learning rate for gradient descent of $\alpha = 10^{-5}$ over 100 iterations.


\begin{figure}
    \centering
    \includegraphics[width=0.7\linewidth]{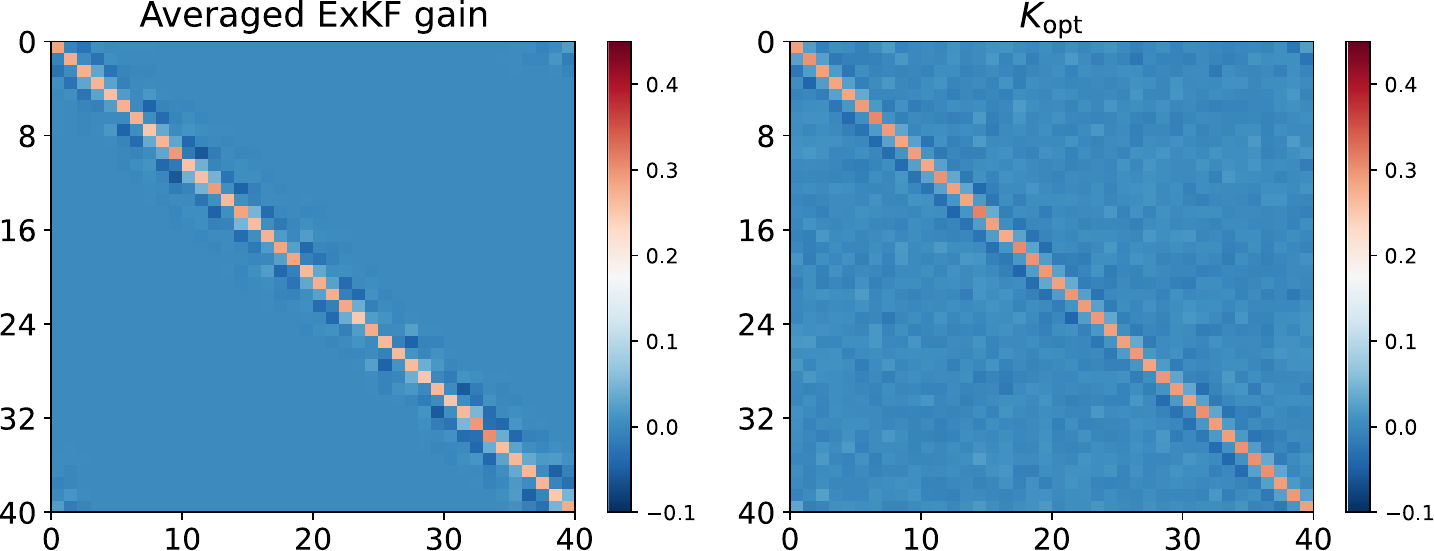}
    \caption{Gains for the Lorenz '96 system. On the left is the extended Kalman filter (ExKF) gain, averaged over 50 iterations. On the right is the learned gain.}
    \label{fig:nonlinear_gain}
\end{figure}

\Cref{fig:nonlinear_gain} shows a comparison of the average of the gain
from the extended Kalman filter for this system with the gain learned by our method. Since we learn a fixed gain and the extended Kalman filter gain is time-varying, and since the extended Kalman filter is only an approximate filter, we do not expect the gains to match. Nonetheless, both methods exhibit gain matrices with a banded and cyclic structure, which is expected due to the correlation structure of states on a circular ring in the Lorenz '96 model. Moreover, due to the symmetry of the model, the optimal fixed gain should be isotropic, which is indeed seen here.

Since we train on a finite set of observations $Y^\dagger_J$, there is the possibility of overfitting and hence that the learned algorithm might not work well on different data sets derived from the same model; see \cite{brocker_sensitivity_2012,mallia-parfitt_assessing_2016} for discussion of out-of-sample error in data assimilation. For this experiment, we use the learned gain for filtering a different trajectory of the same system. This gain still results in a stable filter, and we find the in-sample root-mean-square error (RMSE) to be 0.536 and the out-of-sample RMSE to be 0.562. While the out-of-sample RMSE is slightly higher, the algorithm with the learned gain is still performing well given that the standard deviation of the observation noise is 1.

\subsection{Learning Inflation and Localization for EnKF}
\label{ssec:I&L}

In this section, we test the learning of inflation and localization parameters for an EnKF, using the offline method. \Cref{sec:enkf} provides details of the EnKF method we use. Inflation and localization are both ways of accounting for sampling error 
arising in the EnKF. We discuss the methodologies briefly but point the reader to the review papers \cite{houtekamer_review_2016,carrassi_data_2018} for further details. 

Sampling error leads to systematic underestimation of covariance matrices in EnKFs \cite{sacher_sampling_2008}. A simple way of accounting for this is by choosing a scalar $\lambda > 1$ and scaling (``inflating'') the forecast ensemble spread prior to assimilating observations. Localization, based on the idea of spatial decay of correlation, damps elements of the covariance matrix based on their distance in the physical domain. This reduces the impact of spurious correlations. We specifically use the following form of localization matrix:
\begin{align*}
    (L)_{ik} &= e^{-D_{ik}^2/\ell},
\end{align*}
where $D_{ik}$ is the distance between the variables $i$ and $k$ and $\ell$ is the localization length scale. We take the inflation factor $\lambda$ and localization length scale $\ell$ as unknown parameters, i.e., $\theta = [\lambda, \ell]^\top$.

We consider both the Lorenz '96 model and the Kuramoto--Sivashinsky equation in this experiment. Since these are only two-dimensional optimization problems, we present the contours of the objective function rather than explicitly carrying out gradient descent.

\subsubsection{The Lorenz '96 Model}

Recall, from \cref{ssec:samples}, that we fit the best Gaussian distribution to the ensemble before computing the KL divergence needed as part of the definition of our loss function. However, the covariance matrices are rank deficient whenever the ensemble size $N < D$, preventing use of the formula for the KL divergence for Gaussians \cref{eq:kl_gaussian}. To remedy this rank deficiency, we apply Ledoit--Wolf shrinkage \cite{ledoit_well-conditioned_2004} of the covariance matrices by regularizing them towards the identity matrix as $C\to (1 - \gamma)C + \gamma I$ with the shrinkage parameter  $\gamma=\frac{1}{10}.$

\begin{figure}
    \centering
    \begin{subfigure}{0.45\linewidth}
        \centering
        \includegraphics[width=\linewidth]{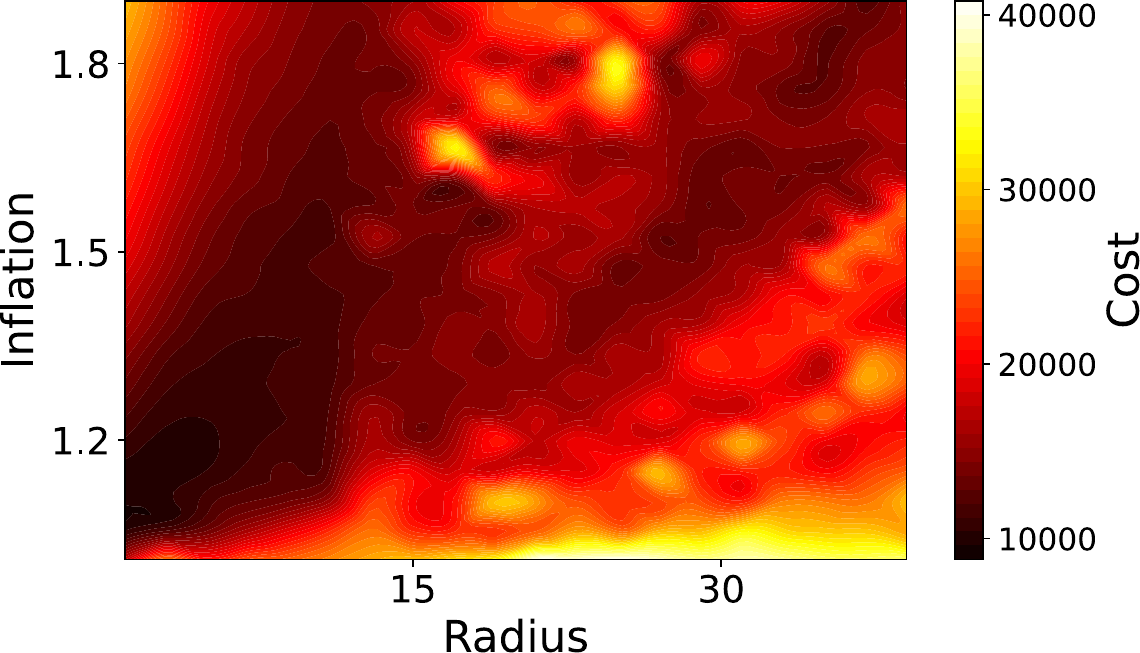}
        \caption{Ensemble size 5}
    \end{subfigure}\hfill\begin{subfigure}{0.45\linewidth}
        \centering
        \includegraphics[width=\linewidth]{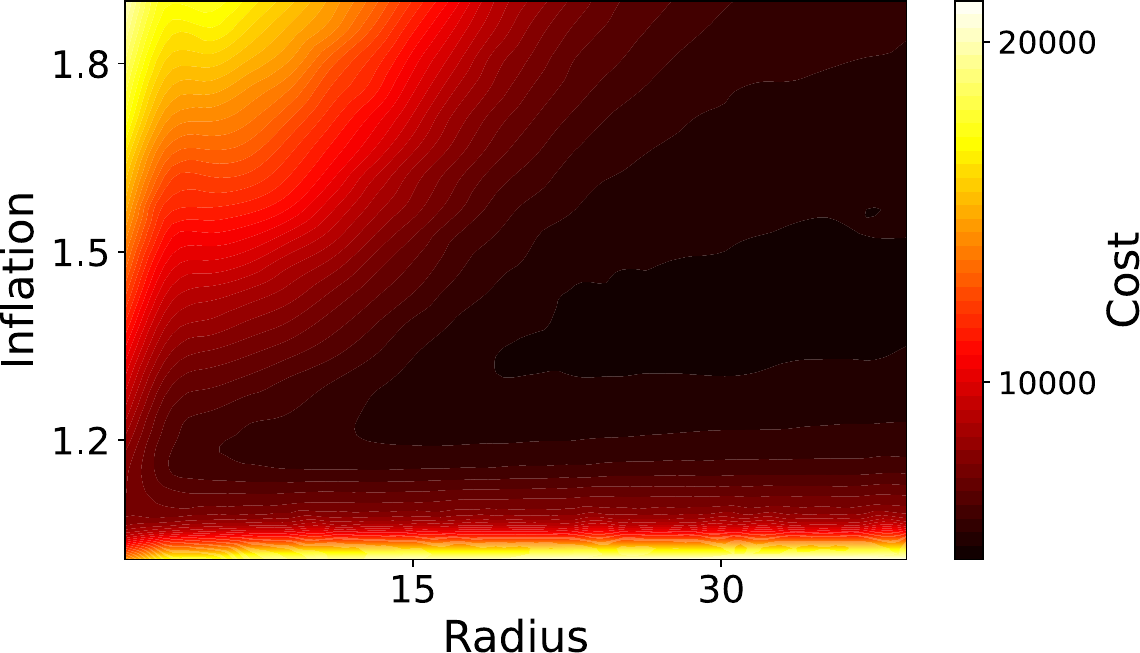}
        \caption{Ensemble size 20}
    \end{subfigure}
    \caption{Contours for the cost function as a function of localization radius and inflation with the Lorenz '96 system.}
    \label{fig:infl_loc}
\end{figure}

\Cref{fig:infl_loc} shows contours of the cost function as a function of inflation and localization radius, for ensemble sizes $N$ of 5 and 20 (both less than $D=40$). In both cases, a larger localization radius generally results in a larger optimal inflation. This is expected because a larger localization radius leads to a larger sampling error, which can be compensated for by inflation. For both cases, the optimal inflation is always greater than 1. This is expected from the systematic underestimation of analysis covariance matrix due to sampling error~\cite{sacher_sampling_2008}.

For ensemble size $N = 5$, a localization radius smaller than 5 and inflation between 1.1 and 1.2 result in the lowest cost. With a larger ensemble size, the sampling error in the covariance matrices is lessened, allowing for a less severe localization. Ensemble size 20 thus achieves the lowest cost with a localization radius greater than 20 and inflation between 1.3 and 1.5. Note also that the lowest costs in the contour plot for the 5-member ensemble are higher than those for the 20-member ensemble, indicating that the EnKF gets closer to the true filter with the larger ensemble size.

\begin{figure}
    \centering
    \begin{subfigure}{0.45\linewidth}
        \centering
        \includegraphics[width=\linewidth]{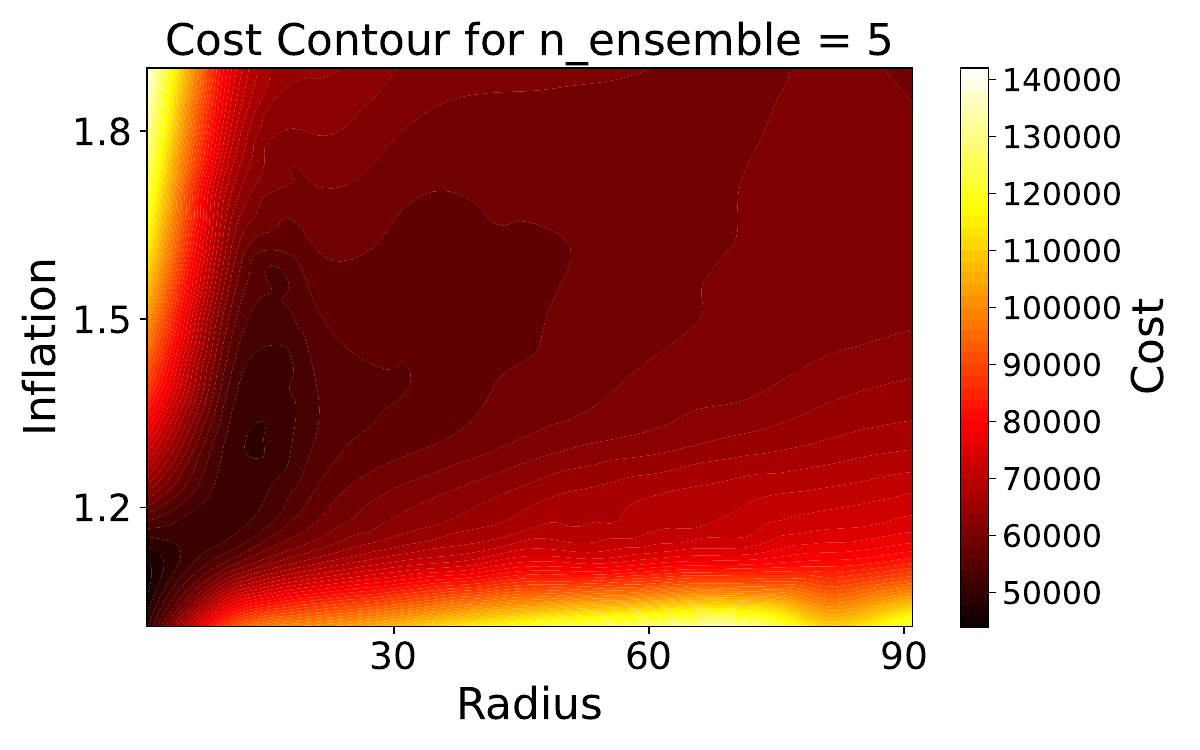}
        \caption{Ensemble size 5}
    \end{subfigure}\hfill\begin{subfigure}{0.45\linewidth}
        \centering
        \includegraphics[width=\linewidth]{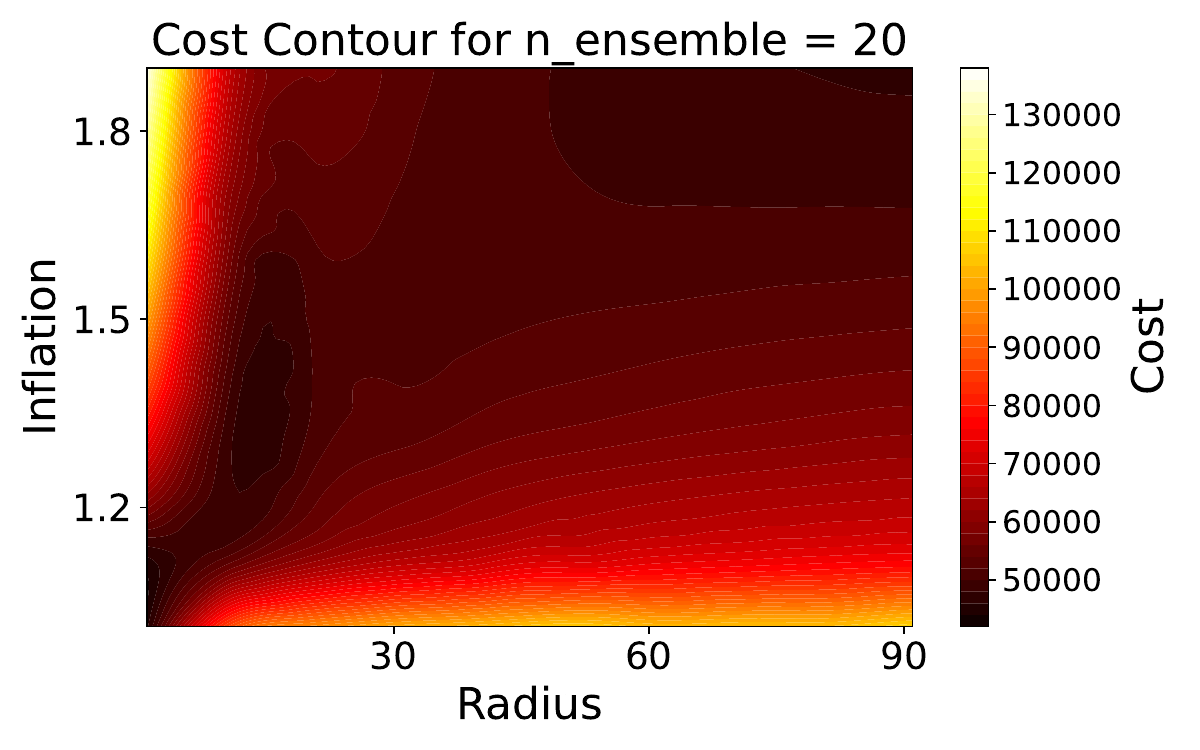}
        \caption{Ensemble size 20}
    \end{subfigure}
    \caption{Contours for the cost function as a function of localization radius and inflation with the Kuramoto--Sivashinsky system.}
    \label{fig:ks_infl_loc}
\end{figure}

\subsubsection{The Kuramoto--Sivashinsky Equation}

\Cref{fig:ks_infl_loc} shows the contours of the cost function 
$\mathsf{J}$, with the Gaussian approximation described in \cref{ssec:samples}, for ensemble sizes $N$ of 5 and 20, with the Kuramoto--Sivashinsky model; recall that we use $D=256$ and so, as in the preceding subsection, Ledoit--Wolf shrinkage is employed for the covariance matrix. $J = 200$ is used for these experiments,
again with the shrinkage parameter set at $\gamma=\frac{1}{10}$.

These plots exhibit similar features to those seen for the Lorenz '96 model. For the 20-member ensemble there appears to be an additional local minimum in the cost function for large radius and large inflation that is not present in the 5-member ensemble. Both ensemble sizes have two local minima for small radius: one with very small inflation and radius, and another with larger inflation and somewhat larger radius.

\section{Conclusions}\label{sec:conclusions}

We introduce a framework for learning or evaluating parameterized filters using variational inference. We apply our computational methodology to learn a Kalman gain in Gaussian filters for both linear and nonlinear dynamical systems, as well as to learn inflation and localization parameters for an ensemble Kalman filter (EnKF).

Future work will investigate learning the parameters of more general filters by approximating the analysis map directly using parameterizations of increasingly complex functions. For instance, $\theta$ can be taken to be the weights of a neural network that parameterize an analysis map that transform a forecast ensemble and observations into an analysis ensemble. Furthermore, state-dependent gains, such as those that appear in the feedback particle filter \cite{yang_feedback_2013}, can be considered.

For the EnKF-based algorithms in this work, the empirical analysis distributions were projected back into the space of Gaussian distributions in order to estimate the KL divergence; see \cref{ssec:samples}. This projection is challenging in high-dimensional settings due both to the statistical estimation error in computing high-dimensional moments and to the computational burden of computing and storing covariance matrices. Future work will consider alternative sample-based objective functions that do not require density estimation, such as those based on strictly proper scoring rules.

\section*{Acknowledgments}

An earlier version of this paper was presented at the Machine Learning for Earth System Modeling Workshop at ICML 2024 \cite{luk_learning_2024}.


\bibliographystyle{siamplain}
\bibliography{references}

\appendix

\section{Filtering Algorithms}\label{sec:filters}

\subsection{Linear 3DVar Filter}\label{sec:gain_appendix}
Recall the dynamics--observation model implied by \eqref{eq:sdm}, \eqref{eq:dm}
in the case where $\Psi(\cdot)=A\cdot$ and $h(\cdot)=H\cdot$ are linear.
We consider a random vector $v_j$ whose law will define the approximate filter:
\begin{subequations}\label{eq:3dvar_linear}
\begin{align}
    \widehat{v}_{j+1} &= Av_{j} + \xi_j,\\
    v_{j+1} &= \widehat{v}_{j+1} + K(y^\dagger_{j+1} + \eta_{j+1} - H\widehat{v}_{j+1}),\\
    v_0\sim &\mathcal{N}(m_0, C_0),\quad \xi_j \sim \mathcal{N}(0, \Sigma) \: \text{i.i.d.},\\
    \eta_{j+1}&\sim\mathcal{N}(0, \Gamma) \: \text{i.i.d.}
\end{align}
\end{subequations}
We proceed to show that the corresponding mean and covariance equations are given by \cref{eq:kalman_fixed_cov}. 

Since $v^\dagger_0$ and $v_0$ are both distributed according to $\mathcal{N}(m_0, C_0)$, $\mathbb{E}[v_0] = \mathbb{E}[v_0^\dagger]$. Then, under the recursions \cref{eq:sdm} and \cref{eq:3dvar_linear}, it follows by induction that $\mathbb{E}[v_j] = \mathbb{E}[v_j^\dagger]$ for all $j\geq 0$. This implies that $\mathbb{E}[y^\dagger_j] = H\mathbb{E}[v_j]$ and that $\mathbb{E}[(y^\dagger_j - H\mathbb{E}[v_j])(y^\dagger_j - H\mathbb{E}[v_j])^\top] = \Gamma$.
The formulae for $\widehat{m}_j = \mathbb{E}[\widehat{v}_j]$ and $m_j = \mathbb{E}[v_j]$ follow immediately. For the covariance $\widehat{C}_{j} = \mathbb{E}[(\widehat{v}_{j} - \widehat{m}_j) (\widehat{v}_{j} - \widehat{m}_j)^\top]$ we have
\begin{align*}
    \widehat{C}_{j+1} &= \mathbb{E}[(A({v}_{j} - {m}_{j}) + \xi_{j+1})(A({v}_{j} - {m}_{j}) + \xi_{j+1})^\top],\\
    &= A C_j A^\top + \Sigma.
\end{align*}
For ${C}_{j} = \mathbb{E}[({v}_{j} - {m}_j)({v}_{j} - {m}_j)^\top]$ we have
\begin{align*}
    {C}_{j+1} &= \mathbb{E}[((I - KH)(\widehat{v}_{j+1} - \widehat{m}_{j+1}) + K\eta_{j+1})((I - KH)(\widehat{v}_{j+1} - \widehat{m}_{j+1}) + K\eta_{j+1})^\top],\\
    &= (I - KH)\widehat{C}_{j+1}(I - KH)^\top + K\Gamma K^\top.
\end{align*}
We note that this covariance update equation for $C_j$ coincides with the equation for the error covariance matrix $\mathbb{E}[(v^\dagger_j - v_j)(v^\dagger_j - v_j)^\top]$, and is known as the Joseph formula \cite{simon_optimal_2006}.



\subsection{Kalman Filter}\label{sec:kf}

Suppose again that $\Psi(\cdot) = A \cdot$ and $h(\cdot) = H \cdot$. Then the solution to the filtering problem is a Gaussian distribution for all time. The update rules for the mean and covariance of the filter are given by the \emph{Kalman filter}~\cite{kalman_new_1960}: the mean updates are given by
\begin{subequations}\label{eq:kalman_mean}
\begin{align}
\widehat{m}_{j+1} &= A m_j,\\ 
m_{j+1} &= \widehat{m}_{j+1} + K_j \bigl(y^\dagger_{j+1}-H\widehat{m}_{j+1}\bigr),
\end{align}
\end{subequations}
where the \emph{Kalman gain}\index{Kalman gain} $K_j$ is determined by the
update rule for the covariances
\begin{subequations}
\begin{align}
\label{eq:udc}
\widehat{C}_{j+1} &= A C_{j} A^\top + \Sigma,\\
K_{j+1} &= \widehat{C}_{j+1} H^\top \bigl(H \widehat{C}_{j+1} H^\top + \Gamma\bigr)^{-1},\\
C_{j+1} &= (I - K_{j+1}H)\widehat{C}_{j+1}.
\end{align}
\end{subequations}

We note that if
the covariance is in steady state then $C_{j+1}=C_j=C_{\text{steady}}$ and 
$\widehat{C}_{j+1}=\widehat{C}_j=\widehat{C}_{\text{steady}}.$ The steady-state covariance and gain are given by solving the equations
\begin{subequations}
\label{eq:ss_kalman}
\begin{align}
    \widehat{C}_{\text{steady}} &= A (I - K_{{\text{steady}}}H)\widehat{C}_{{\text{steady}}} A^\top + \Sigma,\\
    K_{\text{steady}} &= \widehat{C}_{\text{steady}} H^\top (H \widehat{C}_{\text{steady}} H^\top + \Gamma)^{-1}\label{eq:ss_kalman_gain}
\end{align}
\end{subequations}
for the pair $(\widehat{C}_{\text{steady}}, K_{{\text{steady}}})$, and setting
\begin{equation}
\label{eq:ss_kalman2}
C_{{\text{steady}}} = (I - K_{{\text{steady}}}H)\widehat{C}_{{\text{steady}}}.
\end{equation}
Conditions under which the Kalman filter converges to this steady state are given in \cite{lancaster_algebraic_1995}.

\subsection{Ensemble Kalman Filter}\label{sec:enkf}

The \emph{ensemble Kalman filter} \cite{evensen_sequential_1994} resembles the Kalman filter, but operates by evolving an ensemble from which means and covariances are estimated using Monte Carlo sampling; the methodology allows for nonlinear dynamics and observation operators. Here we consider an ensemble square-root filter described in \cite{sakov_relation_2011}, designed in the setting of a linear observation operator $h(\cdot) = H\cdot$. The update at each step for the ensemble matrix $E_j \in \mathbb{R}^{d  \times N}$, whose columns contain the $N$ ensemble members, is given by
\begin{subequations}
\begin{align}
& \widetilde{E}^{(n)}_{j+1} = \Psi({E}_{j}^{(n)}),\quad n=1,\ldots,N,\\
& \widehat{m}_{j+1} = \frac{1}{N}\sum_{n=1}^N \widetilde{E}^{(n)}_{j+1},\\
& \widehat{E}_{j+1} = \lambda (\widetilde{E}_{j+1} - \widehat{m}_{j+1}{1}^\top) + \widehat{m}_{j+1} {1}^\top,\label{eq:inflation}
\\
& \widehat{C}_{j+1} = L\circ\left(\frac{1}{N-1}(\widehat{E}_{j+1} - \widehat{m}_{j+1} {1}^\top)(\widehat{E}_{j+1} - \widehat{m}_{j+1} {1}^\top)^\top\right) + \Sigma,\\
& K_{j+1} = \widehat{C}_{j+1}{H}^\top ({H}\widehat{C}_{j+1}{H}^\top + \Gamma)^{-1},\\
& {m}_{j+1} = \widehat{m}_{j+1} + K_{j+1}(y^\dagger_{j+1} - H\widehat{m}_{j+1}),\\
& {E}_{j+1} = m_{j+1} {1}^\top + \lambda({I} - {K}{H})^{1/2}(\widehat{E}_{j+1} - \widehat{m}_{j+1} {1}^\top).
\end{align}
\end{subequations}
Here $E_j^{(n)}$ is the $n$th column of the ensemble matrix $E_j$, $\lambda$ and $L$ are the inflation factor and localization matrix, respectively (described in \cref{ssec:I&L}); $1$ is a vector of ones; and $\circ$ is the Hadamard, or element-wise, product.
The approximate filtering distribution at time $j$ is given by the empirical measure of the ensemble. That is,
\begin{equation}
    \Pi_{j}^\text{EnKF} = \frac{1}{N}\sum_{n=1}^N \delta_{{E}_{j}^{(n)}}.
\end{equation}

\end{document}

%% file: ex_shared.tex
\usepackage{subcaption}
\usepackage{amsfonts}
\usepackage{graphicx}
\usepackage{epstopdf}
\usepackage{algorithmic}
\ifpdf
  \DeclareGraphicsExtensions{.eps,.pdf,.png,.jpg}
\else
  \DeclareGraphicsExtensions{.eps}
\fi

\usepackage{enumitem}
\setlist[enumerate]{leftmargin=.5in}
\setlist[itemize]{leftmargin=.5in}

\newsiamremark{remark}{Remark}
\newsiamremark{hypothesis}{Hypothesis}
\crefname{hypothesis}{Hypothesis}{Hypotheses}
\newsiamthm{claim}{Claim}

\headers{Learning Optimal Filters Using Variational Inference}{E. Luk, E. Bach, R. Baptista, and A. Stuart}

\title{Learning Optimal Filters Using Variational Inference\thanks{Submitted to the editors 22 March 2025.
\funding{The work of EB, RB, and AMS is supported by a Department of Defense Vannevar Bush Faculty Fellowship awarded to AMS, and by the SciAI Center, funded by the Office of Naval Research (ONR), under Grant No. N00014-23-1-2729. EB was also supported by the Foster and Coco Stanback Postdoctoral Fellowship and by the Office of Naval Research (Grant No. N00014-23-1-2654).}}}

\author{Eviatar Bach\thanks{California Institute of Technology, Pasadena, CA, USA, University of Reading, UK, and National Centre for Earth Observation, Reading, UK 
  (\email{eviatarbach@protonmail.com}).}
\and Ricardo Baptista\thanks{California Institute of Technology, Pasadena, CA, USA.}
\and Enoch Luk\footnotemark[2] \and Andrew Stuart\footnotemark[2]}

\usepackage{amsopn}
